# Entity Candidate Network for Whole-Aware Named Entity Recognition

Wendong He[1], Yizhen Shao[2], Pingjian Zhang[1]*
[1]School of Software Engineering, South China University of Technology, China
[2]School of Computer Science and Engineering, South China University of Technology, China
{sewdhe, csyzshao}@mail.scut.edu.cn, pjzhang@scut.edu.cn

## Abstract

Named Entity Recognition (NER) is a crucial upstream task in Natural Language Processing (NLP). Traditional tag scheme approaches offer a single recognition that does not meet the needs of many downstream tasks such as coreference resolution. Meanwhile, Tag scheme approaches ignore the continuity of entities. Inspired by one-stage object detection models in computer vision (CV), this paper proposes a new no-tag scheme, the Whole-Aware Detection, which makes NER an object detection task. Meanwhile, this paper presents a novel model, Entity Candidate Network (ECNet), and a specific convolution network, Adaptive Context Convolution Network (ACCN), to fuse multi-scale contexts and encode entity information at each position. ECNet identifies the full span of a named entity and its type at each position based on Entity Loss. Furthermore, ECNet is regulable between the highest precision and the highest recall, while the tag scheme approaches are not. Experimental results on the CoNLL 2003 English dataset and the WNUT 2017 dataset show that ECNet outperforms other previous state-of-the-art methods.

## 1 Introduction

NER task plays an essential role in various downstream NLP tasks, including question answering, relation extraction, coreference resolution, etc. Moreover, the quality of entities is a decisive factor. Along the time, researchers treat flat NER as a tagging task and propose many classical approaches. Lample et al. (2016) report an excellent NER performance by combining Long-Short-Term Memory (LSTM, Hochreiter and Schmidhuber 1997), with the Conditional Random Field (CRF, Lafferty et al., 2001). Based on the BIO or BIOES tag scheme, sequential tagging models need to learn different patterns of entities. Moreover, producing a single recognition may lead to a higher error rate when integrating the NER module into other tasks.

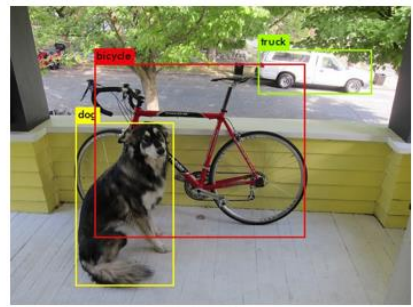

Figure 1: A sample of object detection task. We can view named entities as objects embedded in a textual picture with "bounding boxes."

Different from the tag scheme, human beings determine entities by their start positions and end positions after finishing reading, which is similar to locate objects in a picture. In the field of CV, object detection models detect the boundaries and categories of objects from a picture. Researchers have proposed one-stage models and two-stage models to detect objects. In one-stage models, such as YOLO (Redmon et al., 2015), SSD (Liu et al., 2016) etc., Convolutional Neural Networks (CNNs) are widely used to extract local features.

Figure 1 illustrates the situation where we view entities as objects embedded in textual pictures, and it is more similar to the way that humans extract information than tagging each word. We take this idea one step further in NER task: an enhanced representation suggests enough information to detect the whole entity at each position it occupied. Sequential models are more suitable for long-short term dependence, but detecting entity at each position requires local semantic information. To solve this problem, ECNet first uses multi-head-attention layers to capture global dependency, then ACCN is applied to obtain adaptive context local information via multi-scale receptive fields. Finally, ECNet determines entities by detecting boundaries and classifying categories at each position. In this way, ECNet proposes candidate entities that promise a stronger error-tolerance than traditional methods in both NER and downstream tasks. In the Whole-Aware Detection, we treat non-entity as an equal class which leads to unbalance in samples. To resolve this issue, we adopt the Entity Loss to jointly train ECNet on both location task and classification task in an end-to-end manner.

In summary, the main contributions of our work are as follows:

- This paper views NER as an object detection task and proposes an intuitive no-tag scheme, the Whole-Aware Detection. In this way, NER models are highly regulable between precision and recall, and offer more candidate entities for downstream tasks;

- We propose the Entity Candidate Network to detect the entity, and we use an Entity Loss to measure each candidate. Moreover, we design a new convolutional layer, Adaptive Context Convolution Network, to extract, adjust and aggregate hierarchical information;

- ECNet does not contain any sequence layers or CRF layers in its core architecture. Meanwhile, we do not finetune the pre-trained word embedding. However, we provide a trainable pattern embedding to make up mode information of entities. These make ECNet more efficient on GPU. Besides, ECNet is small and easy to be integrated into downstream tasks.

## 2 Related Work

Generally speaking, the study of NER falls into two categories: the tag scheme approaches and the no-tag scheme approaches. The tag scheme approaches treat NER as a sequential tagging task and usually adopt LSTM as the backbone network. Moreover, techniques like CRF (Lafferty et al., 2001) are often utilized to promote the accuracy of prediction. However, the tag scheme approaches suffer from the complexity of the entity pattern and lead to low precision. Recently no-tag scheme approaches have attracted interests due to their generalization ability in both flat and nest NER tasks. The significant difficulty of no-tag scheme approaches is how to determine the boundary of each entity. MGNER proposed by Xia et al. (2018) utilizes a span-based scheme to determine whether a span is an entity in both flat and nest NER tasks. ARNs proposed by Lin et al. (2019) captures anchor words and then find their boundaries. No-tag scheme methods can handle the complexity of the entity patterns but still face the difficulty of determining the boundary of an entity.

Pre-trained embedding from language models is added to many NER models to obtain contextual representations and improve performance furthermore. Peters et al. (2018) proposed a deep language model, ELMo, which is trained on large corpus and generates dynamic contextual features for words and combines bidirectional states to capture context-dependent semantic information.

Bahdanau et al. (2014) introduced the attention mechanism to machine translation. The attention mechanism proves its potential in bidirectional sequence modeling, becoming a significant part of various NLP tasks. Vaswani et al. (2017) proposed a multi-head-attention mechanism to capture global dependencies across the entire sentence regardless of the distance between different words, achieving significant success in machine translation and language modeling.

When it comes to calculating the N-gram information, CNN is capable of encoding the local information, achieving excellent results in many NLP tasks. Yoon Kim (2014) proposed the TextCNN for sentence classification, and Zhang et al. (2016) leveraged a character-level CNN to represent spelling characteristics. Dauphin et al. (2016) proposed Gated Linear Unit (GLU) to enhance CNN with a gating mechanism, and exploited it to adjust information, making a breakthrough in machine translation. The size of the receptive field is a subtle factor of high-quality local information. The larger receptive field covers a broader range of sentences but may lead to the loss of local information. In the field of CV, Li et al. (2019) proposed SKNet to adjust the receptive field by a selection mechanism adaptively. As for NER task, Chen et al. (2019) used particular relation layers to fuse hybrid local information extracted by CNNs to generate global text representation.

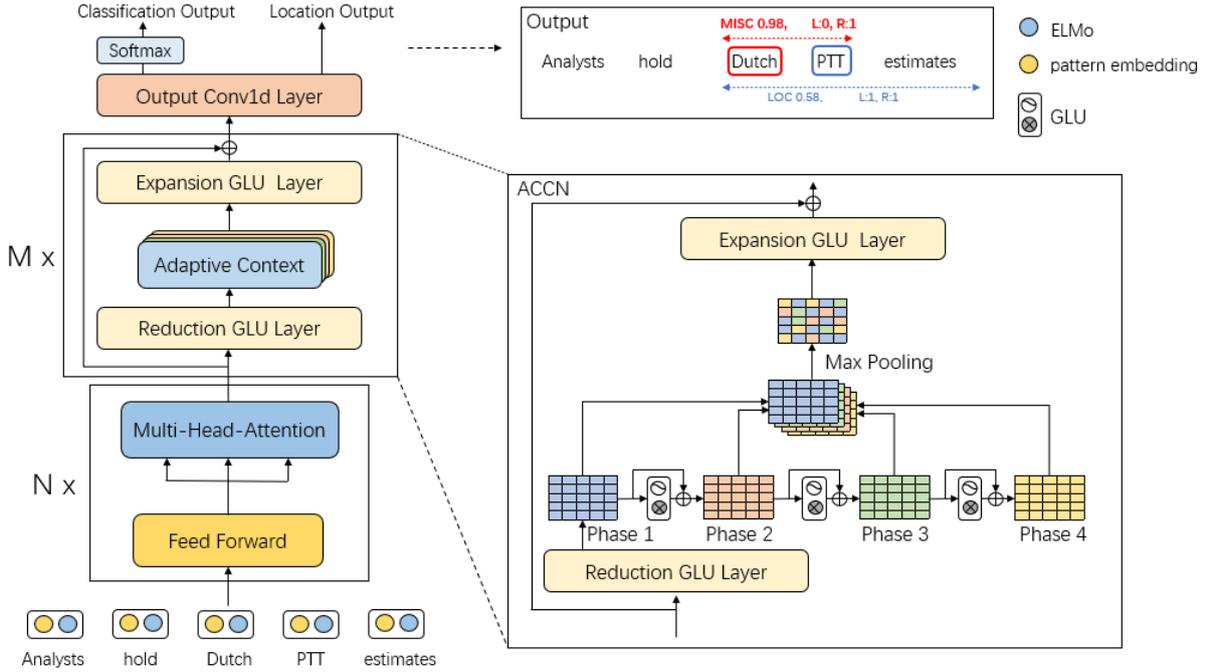

Figure 2: The Whole-Aware Detection is executed at each position. ECNet treats each word of the entity "*Dutch PTT*" as a candidate, and ECNet obtains the correct span and type of the entity via multiple candidates.

## 3 The Framework

Here we introduce the Whole-Aware Detection and the details of ECNet. Tag scheme approaches ignore the continuity of entities and lack of generalization in nest NER task. However, entities are shift-invariant and possess continuity, which is worthy of concern. Different from recent no-tag scheme approaches, ECNet collects multi-scale context information and does not need to pay too much attention to handling the entity boundary. In this manner, ECNet omits duplicate computations and produces multiple candidate entities for better decisions. Figure 2 illustrates the detail of ECNet and the form of the output.

ECNet receives a sentence $s$ with $n$ words $s = \{x_1, ..., x_n\}$. We adopt ELMo (Peters et al., 2018) to generate contextual representation. We do not finetune ELMo. Instead, we define nine word-patterns to supplement the mode information of the entity by the pattern embedding $\text{PAT}(x_i)$ as the following description:

a) Uppercase letters only like "*ABCDE*";

b) Lowercase letters only like "*abcde*";

c) Letters only, the first letter being uppercase and others being lowercase like "*Abcde*";

d) Letters only, the first letter being lowercase, or the first letter being uppercase letter followed by other uppercase letters like "*aBCDE*" and "*AbCDe*";

e) Uppercase letters, punctuations, and numbers like "*ABCDE12#*";

f) Lowercase letters, punctuations, and numbers like "*abcde12#*";

g) Letters, punctuations and numbers, letters obeying pattern c) like "*Abcde12#*";

h) Letters, punctuations and numbers, letters obeying pattern d) like "*AbCDe12#*";

i) Punctuations and numbers only like "*1#2^*".

ECNet concatenates ELMo with the pattern embedding as the input of each word $x_i$. The multi-head-attention (Vaswani et al., 2017) is a dot-product self-attention, and ECNet applies $N$ identical multi-head-attention layers, denoted as MHA, to capture global dependency. We add unique position information following Vaswani et al. (2017). Dropout (Hinton et al., 2012) and layer normalization (Ba et al., 2016) are adopted to prevent overfitting:

$$e_i = [\text{ELMo}(x_i); \text{PAT}(x_i)] \quad (1)$$

$$\{h_1, h_2, \dots, h_n\} = \text{MHA}(\{e_1, e_2, \dots, e_n\}) \quad (2)$$

Since global dependency encodes entity information into some words related by entities, like pronouns, which may be detected as entities incorrectly. In order to solve the above problem and achieve Whole-Aware Detection, we need to enhance entity representation by correct local information, like N-gram information. CNN can capture N-gram information, but each word may not be of equal concern. For example, in the tri-gram "*Dutch PTT estimates*," we mainly need the information of the first two words.

Inspired by ResNet (He et al., 2016) and Res2Net (Gao et al., 2019) in the field CV, we propose the Adaptive Context Convolution Network (ACCN), to exploit local information and encode entity information into each position. ACCN adopts GLU (Dauphin et al. 2016), a CNN with a gating mechanism, to reduce the weight of irrelevant words and obtain the correct semantic information. Thus, ACCN can adaptively choose the context in a convolution window of fixed size. However, information from windows of fixed size is insufficient to extract entity, but directly fusing information from CNN of different granularity will lose some local information. We propose a soft information fusion mechanism, the so-called Multi-Phase Semantic Fusion, which adopts a fixed receptive field to cover entity information of different granularity. Finally, ACCN fuses multi-scale contexts and selects the most important aspects across different channels.

ECNet comprises $M$ identical ACCN layers. Each ACCN layer conducts the following operation:

1) ACCN reduces the input dimension via a reduction-factor $rd$ by the GLU layer with the kernel of size 1;

2) ACCN performs the Multi-Phase Semantic Fusion between three GLU layers with the kernel of size 3. In Multi-Phase Semantic Fusion, the input feature map is skip-connected with the output feature map to produce the output. We carry out the above operation three times and obtain four feature maps with multi-scale context information of different granularity. ACCN obtains semantic structure information of various granularity when we stack it;

3) ACCN juxtaposes the above four feature maps together and employs the channel-max-pooling operation to select the strongest signal at each position among channels. In this operation, ACCN fuses the multi-scale information and keeps the most important aspects;

4) ACCN expands the representation to the original dimension by GLU layer with the kernel of size 1;

5) The original input is added to the output.

Zero-padding masking and the ELU activation function (Clevert et al., 2015) are following to keep uniform shapes and prevent noise from artifact in a batch.

Here are the equations of $j$-th ACCN layer where $j \epsilon \{1, \dots, M\}$, $k$ is the kernel size, $\otimes$ is the element-wise multiplication, $\sigma$ is the sigmoid function, and $p$ denotes different phases in Figure 2. $h_i^j$ represents the representation of $x_i$ in $j$-th ACCN layer. $\text{GLU}^{j^{preduce/expand/p}}$ represents the reduction GLU layer, the expansion GLU layer, and the different GLU layers in $p$-th phase individually in $j$-th ACCN layer:

$$\begin{aligned}\text{GLU}(h_i^{j-1}, k) = \text{conv}_a(h_i^{j-1}, k) \\ \otimes \sigma(\text{conv}_b(h_i^{j-1}, k))\end{aligned} \quad (3)$$

$$c_i^p = \begin{cases} \text{GLU}^{j^{reduce}}(h_i^{j-1}, 1), & p = 0 \\ \text{ELU}\left(\text{GLU}^{j^p}(c_i^{p-1}, 3)\right) + c_i^{p-1}, & p = 1,2,3 \end{cases} \quad (4)$$

$$\tilde{c}_i^j = \text{maxpooling}([c_i^0; c_i^1; c_i^2; c_i^3]) \quad (5)$$

$$o_i^j = \text{ELU}\left(\text{GLU}^{j^{expand}}(\tilde{c}_i^j, 1)\right) \quad (6)$$

$$h_i^j = \text{ELU}(o_i^j + h_i^{j-1}) \quad (7)$$

Suppose we have $c$ types of entities and one additional non-entity, we use $c + 1$ convolution filters to generate $\hat{y}_i^l \in [0, 1]$ which represents the confidence of $l$-th type where $l \in \{1, ..., c, c + 1\}$:

$$\hat{y}_i^l = \text{Softmax}(\text{ReLU}(\text{conv}_l(h_i^M, k))) \quad (8)$$

We also need the left offset $\hat{L}_i$ and the right offset $\hat{R}_i$ at each position. We employ ReLU activation function (Nair and Hinton, 2010) to avoid negative results:

$$\hat{L}_i = \text{ReLU}(\text{conv}_{left}(h_i^M, k)) \quad (9)$$

$$\hat{R}_i = \text{ReLU}(\text{conv}_{right}(h_i^M, k)) \quad (10)$$

We use convolution filters with a kernel of size 3. In this manner, ECNet determines a set of entity candidates at all positions. During inference, we only keep candidates with the highest confidence in the overlapping area.

## 4 Loss Function

For a given sentence $s$ with $n$ words: $s = \{x_1, ..., x_n\}$, the golden label for each word $x_i$ expressed as $y_i = (L_i, R_i, y_i^1, ..., y_i^c, y_i^{non})$ when there are $c$ types of entities and one non-entity. The type label $(y_i^1, ..., y_i^c, y_i^{non})$ is a one-hot vector. The offsets are zero when $y_i^{non}$ is 1.

Without tag schemes and the chain-probability loss, ECNet can eliminate impacts from other tags and utilize cohesion information of every span. To this end, we shall define the Entity Loss. ECNet first employs the smooth L1 loss (Ren, 2015) as offset loss. We only calculate offset loss at the correct position. Here is the loss function for the left offset:

$$L_{loc}(L, \hat{L}) = \begin{cases} 0.5(L - \hat{L})^2, & |L - \hat{L}| < 1 \\ |L - \hat{L}| - 0.5, & otherwise \end{cases} \quad (10)$$

The loss function for the right offsets is defined similarly. The boundary loss is then simply the sum of them:

$$L_{bound} = L_{loc}(L, \hat{L}) + L_{loc}(R, \hat{R}) \quad (11)$$

Because of the unbalance in positive and negative samples, we apply focal loss (Lin et al., 2017) as classification loss to adjust the unbalance by the following equations where $\alpha$ and $\gamma$ are regulatory factors to adjust the unbalance:

$$L_{conf}(y, \hat{y}) = \begin{cases} -\alpha(1 - \hat{y})^\gamma \log \hat{y}, & y = 1 \\ (\alpha - 1)\hat{y}^\gamma \log(1 - \hat{y}), & y = 0 \end{cases} \quad (12)$$

We sum the classification loss $L_{conf}$ with the boundary loss $L_{bound}$ as the Entity Loss $L_{entity}$, where $\beta$ is a scaling coefficient, $N$ is the length of

| Dataset | | Train | Dev | Test |
|---|---|---|---|---|
| CoNLL 2003 | Sentence | 14,987 | 3,466 | 3,684 |
| | Entity | 23,499 | 5,942 | 5,648 |
| WNUT 2017 | Sentence | 3,394 | 1,009 | 1,287 |
| | Entity | 3,160 | 1,250 | 1,589 |

Table 1: Statistics of CoNLL2003 dataset and WNUT2017 dataset.

the sentence, and $T$ is the amount of the positions with entities:

$$L_{entity} = \beta \frac{1}{N} \sum L_{conf} + \frac{1}{T} \sum (1 - y^{non}) L_{bound} \quad (13)$$

## 5 Experiment

To verify the advancement of ECNet as well as the Whole-Aware Detection, we experiment on two benchmark datasets: CoNLL2003 dataset (Tjong Kim Sang and De Meulder, 2003) and WNUT2017 dataset (Derczynski et al., 2017). Table 1 shows some statistics of both datasets.

**CoNLL2003**
CoNLL-2003 English NER dataset extracts text with annotations for 4 types of entities: PERSON, LOCATION, ORGANIZATION, and MISC.

**WNUT2017**
WNUT2017 dataset contains a small number of emerging entities and is disturbed by social noise, with annotations for 6 types of entities: PERSON, LOCATION, GROUP, CREATIVE-WORK, PRODUCT, and MISC.

We implement ECNet with Pytorch (Paszke et al., 2017) and AllenNLP library (Gardner at al., 2017). We set up parameters of ECNet, referring to Vaswani et al. (2017), He et al. (2015), and Lin et al. (2017). We remove the sentences only contain one class (non-entity is an equal class) during training in both two datasets, which are discovered to be harmful to build syntax structure.

**Optimization**
We adopt Adam optimizer (Kingma and Ba, 2014) to train ECNet. The batch size is set as 64. The weight decay is set as 1e-4, and the initial learning rate is set as 1e-4. We reduce the learning rate by 0.1 during training at every 100 epochs. We train ECNet with 1000 epochs.

**Word Embedding**
We only use ELMo and our small pattern embedding. The dimension of ELMo is 1024, and the dimension of the pattern embedding is 128. We do not finetune ELMo during training.

| Model | F1 | P/R |
|---|---|---|
| (Chiu and Nichols 2016) | 91.62 | 91.39 / 91.85 |
| (Peters et al. 2018) | 92.22 | NA / NA |
| BERT (Devlin et al., 2018) | 92.40 | NA / NA |
| MGNER (Xia et al., 2018) | 92.28 | NA / NA |
| GRN (Chen et al., 2019) | 92.34 | 92.04 / 92.65 |
| ECNet (threshold = 0.1) | 90.91 | 89.09 / **92.81** |
| ECNet (threshold = 0.7) | **92.71** | **94.11** / 91.35 |

Table 2: Performance comparison on CoNLL2003 with F1, precision, and recall under two thresholds.

| Model | F1 | P/R |
|---|---|---|
| SpinningBytes (Däniken and Cieliebak, 2017) | 40.78 | NA / NA |
| (Aguilar et al., 2018) | 45.55 | 61.06 / 36.33 |
| ECNet (threshold = 0.1) | 44.79 | 52.94 / **38.82** |
| ECNet (threshold = 0.6) | **47.02** | **67.84** / 35.98 |

Table 3: Performance comparison on WNUT2017 with F1, precision, and recall under two thresholds.

**Multi-Head-Attention and Adaptive Context Convolution Network**
ECNet contains 3 multi-head-attention layers and 3 ACCN layers. The dimension of all these layers is 512. We set 8 heads in each multi-head-attention layer and set $rd$ in ACCN as 0.25, referring to Bottleneck (He et al., 2016).

**Loss Hyper-Parameter**
We set $\gamma$ as 2 and $\alpha$ as 0.05 for the focal loss. In order to balance the loss, we set $\beta$ as 10.

## 6 Result Comparison

We compare ECNet with existing state-of-the-art NER approaches and evaluate them based on span-based F1. On testing, we round the fractional part of the left and right boundaries. It is worth saying, exploiting external knowledge to boost performance has been prevalent recently and makes it more robust than the original models. We primarily focus on the comparison between models utilizing external knowledge or pre-trained word embeddings.

**The Result of CoNLL2003**
We compare ECNet with previous state-of-the-art models based on LSTM and CNN. Chiu and Nichols (2016) combined CNN and BiLSTM with word embedding trained on Wikipedia. Peters et al. (2018) first proposed ELMo and integrated it into BiLSTM-CRF NER architecture. We also compare ECNet with GRN proposed by Chen et al. (2019), which adopts ELMo and CNNs for NER on the tag-scheme. As for not-tag scheme approaches, MGNER (Xia et al., 2018) uses different fixed length windows to generate proposal regions and determines whether a span is an entity. BERT-base (Devlin et al., 2018) finetunes NER task on the CoNLL2003 dataset after trained on a large corpus.

The comparison results in Table 2 show that ECNet obtains the best F1 score of 92.71 and outperforms other state-of-the-art methods.

**The Result of WNUT2017**
The comparison between ECNet and previous state-of-the-art models on the WNUT2017 dataset is shown in Table 3. SpinningBytes (Däniken and Cieliebak, 2017) proposed sentence embedding to get social media entities. Aguilar et al. (2019) encoded noisiness with the FastText (Bojanowski et al., 2016) vector by BiLSTM to recognize entities.

As shown in Table 3, ECNet obtains the best F1 score of 47.02 and outperforms other state-of-the-art methods.

## 7 Regulability of ECNet

The tag-scheme approaches tag each sentence only once, which may cause low precision in NER and cascading error in downstream NLP tasks. ECNet provides multiple detections with confidence and representation for each candidate. We can choose an appropriate threshold for different NLP tasks basing on the demand for high precision or high recall. Meanwhile, ECNet gives a priori probability for each candidate and provides local candidate representations for downstream tasks. For example, we can integrate ECNet to the coreference resolution models to avoid the computations of span-based mention prediction, improve the mention quality, and calculate the mention confidence to the final loss.

In the NER task, we analyze the accuracy, recall, and F1 under different thresholds on both datasets in Table 4 and Table 5. The experimental results show that redundant candidates guarantee the highest precision, while under a low threshold, ECNet keeps a noteworthy precision and retains more candidates, which gives more possibilities for downstream tasks. When compared with recent no-tag scheme approaches, ECNet scans the whole entities directly without the needs of proposal regions or repetitive boundary comparisons. Detecting on each word promises a stronger error-tolerance for any possible entity.

| Threshold | F1 | Precision | Recall |
|---|---|---|---|
| 0.1 | 90.91 | 89.09 | **92.81** |
| 0.2 | 91.44 | 90.15 | 92.76 |
| 0.3 | 91.85 | 91.08 | 92.63 |
| 0.4 | 92.06 | 91.74 | 92.38 |
| 0.5 | 92.29 | 92.43 | 92.15 |
| 0.6 | 92.53 | 93.18 | 91.89 |
| 0.7 | **92.71** | 94.11 | 91.35 |
| 0.8 | 92.68 | 94.86 | 90.59 |
| 0.9 | 92.55 | **96.02** | 89.33 |

Table 4: Performance on CoNLL2003 with F1, precision and recall under different thresholds

| Threshold | F1 | Precision | Recall |
|---|---|---|---|
| 0.1 | 44.79 | 52.94 | **38.82** |
| 0.2 | 44.71 | 53.13 | 38.59 |
| 0.3 | 45.09 | 53.99 | 38.71 |
| 0.4 | 45.43 | 56.55 | 37.96 |
| 0.5 | 46.13 | 62.18 | 36.66 |
| 0.6 | **47.02** | 67.84 | 35.98 |
| 0.7 | 45.17 | 69.04 | 33.56 |
| 0.8 | 42.03 | 71.96 | 29.68 |
| 0.9 | 38.45 | **76.59** | 25.67 |

Table 5: Performance on WNUT2017 with F1, precision and recall under different thresholds

Besides, ACCN obtains cohesion information of different granularity, which gives a higher possibility of discovering emerging entities. Although ECNet achieves the best performance, it brings a new problem that the unbalance in positive and negative samples may lower down the recall.

## 8 Ablation Study

In this section, we shall analyze each part of ECNet and explain each case in Table 6. For better comparison with ACCN, we design ResGLU following ResNet (He et al., 2016). Each ResGLU layer comprises three GLU layers. We utilize two GLU layers to reduce and expand the dimension by $rd$, and one GLU layer is to obtain tri-gram information. Figure 3 illustrates an overview of the ResGLU layer.

We wipe out or replace some parts of ECNet for some purpose of ablation study:

1) W/o the pattern embedding: whether or not pattern embedding can supplement pattern information as we do not finetune ELMo (it can get a considerable speed-up);

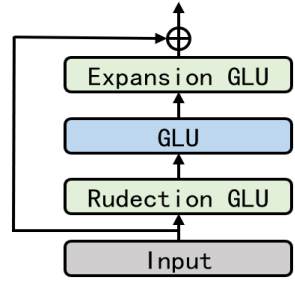

Figure 3: Overview of ResGLU.

| | F1 | |
|---|---|---|
| | CoNLL 2003 | WNUT 2017 |
| w/o pattern embedding | 92.47 | 46.34 |
| w/o ACCN | 87.05 | 39.32 |
| Replaced by ResGLU | 91.36 | 43.15 |
| Larger kernel size | 91.82 | 44.27 |
| **Completed ECNet** | **92.71** | **47.02** |

Table 6: Ablation study on CoNLL2003 and WNUT2017, the completed ECNet gets the best.

2) W/o ACCN layers: whether or not sentence-level encoder can involve local information. For a fair comparison, we set 6 multi-head-attention layers to maintain the size;

3) Replace ACCN layers with ResGLU layers: to show the necessity of ACCN as well as the Multi-Phase Semantic Fusion;

4) Larger kernel size in ACCN: set the kernel size in each ACCN layer as 5.

All ablation models fit the same experimental settings. Table 6 reports the results on both datasets, and the last column shows the performance of the completed ECNet. We analyze each case and come to the following conclusions:

1) The entity has a statistical pattern, and the pattern embedding is capable of distinguishing different word patterns. By the way, if we integrate more language models, ECNet will perform better theoretically;

2) Removing ACCN causes noteworthy performance degradation. The sentence-level encoder cannot collect fine-grained information, and some related words will be detected as entities incorrectly. On the contrary, ACCN can filter out the noise and obtain semantic information of different granularity without the loss of information;

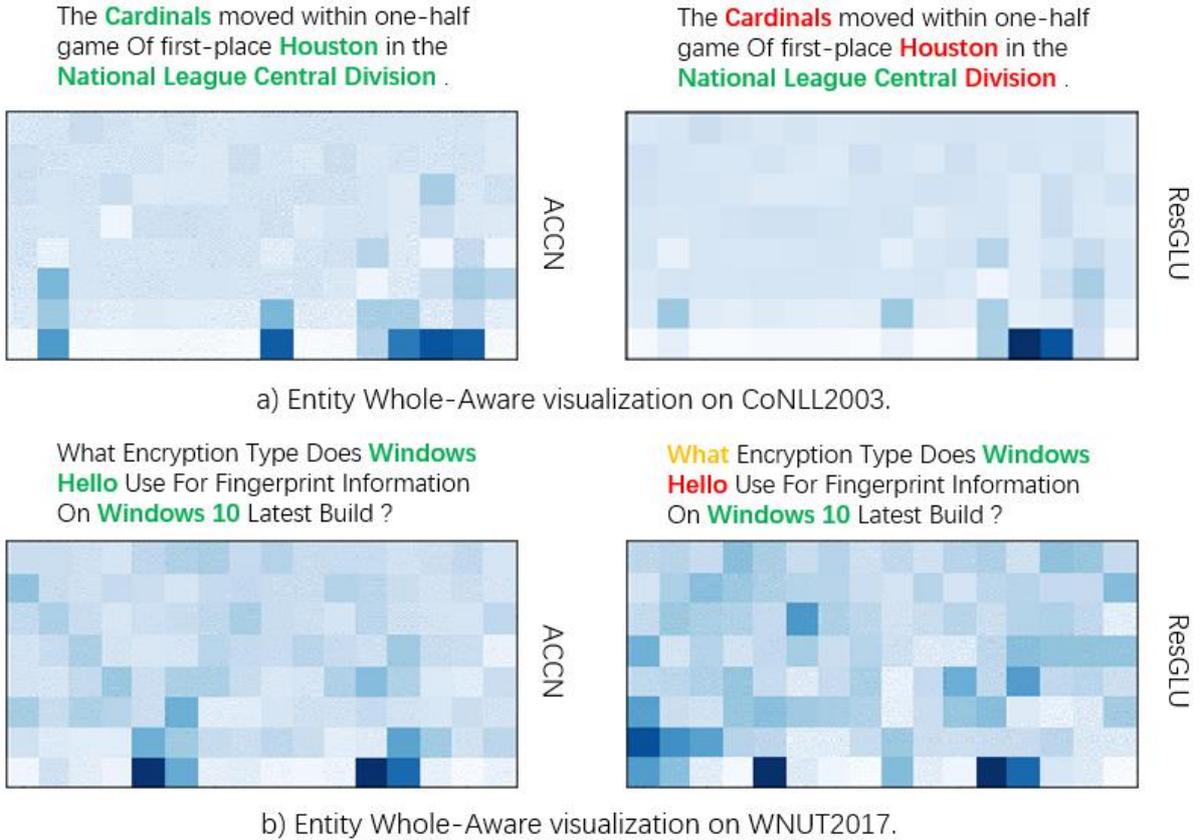

Figure 4: Entity Whole-Aware visualization of ECNet (ACCN versus ResGLU). Words in green are correct recognitions, words in red are missing recognitions, and words in yellow are wrong recognitions.

3) Using ResGLU layers instead will lose some semantic information due to its rigid CNN structure and insufficient context. However, ACCN provides adaptive context information which preserves correct entity information;

4) Violently expanding receptive fields may lose correct local semantic information and weaken the focus on relevant information. However, windows of fixed-length are insufficient. ACCN applies Multi-Phase Semantic Fusion to fuse fine-grained information into sentence-level information and keep the information of different granularity.

## 9  Entity Whole-Aware Visualization

When it comes to Whole-Aware Detection, we need to filter entities from a noisy text space. As shown in Figure 4, we analyze the recognition capability of ACCN versus ResGLU. We select two sentences with the length of 16 from both datasets, intercept feature maps after the above layers, reduce the dimensions to 8 by Principal Components Analysis (PCA) and draw the figures.

Generally speaking, as shown in Figure 4, ACCN layers are more capable of eliminating the harmful impacts of non-entities. In CoNLL2003, "*Cardinals*," "*Houston*," and "*Division*" are detected correctly by ACCN but are ignored by ResGLU. Meanwhile, ACCN eliminates the noise of "*one-half*." In WNUT2017, ResGLU mistakes "*What*" and misses "*Hello*." On the contrary, ACCN makes correct detections.

## 10  Conclusion

In this paper, we presented a promising no-tag scheme to the NER task, the Whole-Aware Detection. To this end, we proposed a novel model, ECNet, with a specially designed network, ACCN. Experimental results show that ECNet can outperform other state-of-the-art methods and provide more choice of entities for downstream NLP tasks. Moreover, we discuss the advantages of Whole-Aware Detection. We hope our work can inspire some tag tasks in NLP.